\documentclass[letterpaper, 10 pt, conference]{ieeeconf} 

\IEEEoverridecommandlockouts                            

\pdfoutput=1

\usepackage{graphicx}
\usepackage{diagbox}
\usepackage{multirow}
\usepackage{siunitx}
\usepackage{stackengine}
\usepackage{comment}
\usepackage[caption=false]{subfig}
\usepackage{hyperref}
\usepackage{color}

\usepackage[symbol]{footmisc}

\newcommand\xrowht[2][0]{\addstackgap[.5\dimexpr#2\relax]{\vphantom{#1}}}
\title{\LARGE \bf
Benchmarking Domain Randomisation for\\ Visual Sim-to-Real Transfer
}

\author{Raghad Alghonaim$^{1}$ and Edward Johns$^{1}$
\thanks{$^{1}$ The Robot Learning Lab at
        Imperial College London
        {\tt\small \{raa318, e.johns\}@imperial.ac.uk}}%
}

\begin{document}

\maketitle
\thispagestyle{empty}
\pagestyle{empty}

\begin{abstract}
Domain randomisation is a very popular method for visual sim-to-real transfer in robotics, due to its simplicity and ability to achieve transfer without any real-world images at all. Nonetheless, a number of design choices must be made to achieve optimal transfer. In this paper, we perform a comprehensive benchmarking study on these different choices, with two key experiments evaluated on a real-world object pose estimation task. First, we study the rendering quality, and find that a small number of high-quality images is superior to a large number of low-quality images. Second, we study the type of randomisation, and find that both distractors and textures are important for generalisation to novel environments.

\end{abstract}

\section{Introduction}

In recent years, deep learning has been successfully applied to a range of robotics applications, with particular success in those which require visual observations for control \cite{levine2016end, johns2021coarse, james2017transferring, DBLP:journals/corr/abs-1709-07857}. Here, convolutional neural networks (CNNs) enable learning of task-specific visual features directly from data, avoiding the need for laborious task-specific engineering, and potentially outperforming methods using hand-crafted visual representations. However, the reliance of deep learning on large labelled datasets presents a significant challenge.

One of the most promising solutions is sim-to-real transfer, where training is performed in simulation, and a controller is transferred directly to the real world. Of the many sim-to-real transfer methods for vision, domain randomisation \cite{tobin2017domain, sadeghi2016cad2rl, james2017transferring} is the most popular, since it is simple to implement, and can achieve zero-shot transfer without any real-world data. However, despite its popularity, there are no significant works which benchmark the different types of domain randomisation for visual sim-to-real transfer. Recent benchmarking work has studied sim-to-real for dynamics \cite{valassakis2020crossing}, but the visual sim-to-real problem is a distinct problem and is typically treated independently from dynamics.

\begin{figure}[tb] 
    \centering
    \includegraphics[width=8cm]{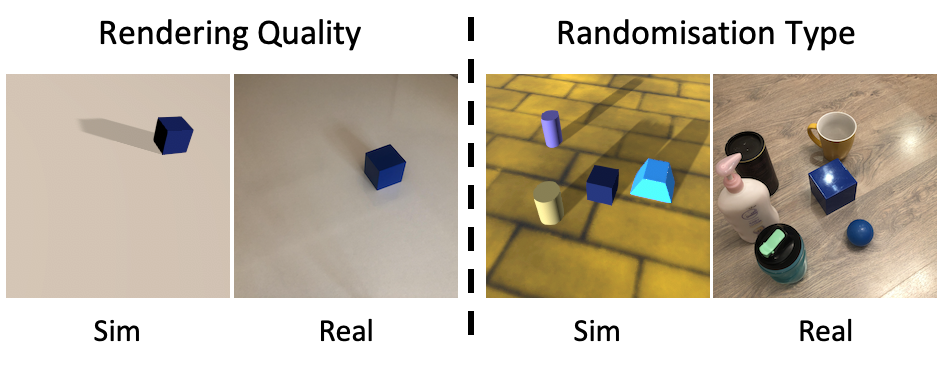}%
    \caption{An overall illustration of our experiments. Left: we study the effect of the rendering quality on the sim-to-real transfer performance with known environment settings. Right: we examine the impact of different randomisation types on the transferability of models trained in simulation and tested in challenging real-world scenarios.}
    \label{fig:overall_illustration}
\end{figure}

In this paper, we study a range of different design choices and empirically evaluate their effect on a real-world task. We divide the paper into two distinct experiments, which study two modes of design choices. In the first experiment, we consider the case when the scene content is known in advance, and the primary role of sim-to-real is to model the effect of illumination and image noise on the observed image. Here, we study how the quality of the rendered images, in terms of the fidelity of graphics pipeline, affects the sim-to-real performance. We also study the optimal trade-off between low-quality and high-quality images, given a fixed amount of rendering time. In the second experiment, we consider the case when the scene content is not known in advance, and the role of sim-to-real is to achieve robustness to the unknown, such as illumination conditions and clutter. Here, we study how different types of randomisation, such as colours and textures, affect the sim-to-real performance.

We evaluated both experiments with a 6D object pose estimation task, with a manually-labelled real-world dataset. This not only evaluates sim-to-real for a pose estimation module within a wider control pipeline, but also is a proxy to end-to-end control methods \cite{james2017transferring} which implicitly localise important objects within an image. Our results highlight the importance of high-quality images for sim-to-real transfer, and the importance of randomising both distractors and textures. Our supplementary video gives an overview of our experimental setup, as well as visualisations of results{\color{red}\footnote[1]{\href{https://www.robot-learning.uk/benchmarking-domain-randomisation}{\color{blue}{www.robot-learning.uk/benchmarking-domain-randomisation}}}}.

\section{Related Work}

Deep neural networks are known to be sample-inefficient, as a massive amount of annotated images is required for the training process to succeed. Due to the expensive data collection on real robots \cite{levine2018learning}, researchers have shifted towards using simulators to generate the training images. Nonetheless, models trained with synthetic images only are not capable of directly transferring to the real world \cite{tremblay2018training}, because of their various physical and visual differences, a problem referred to as the reality gap. Sim-to-real transfer is an active area of research to bridge the gap with two main approaches, \textit{Domain Adaptation} and \textit{Domain Randomisation}. To close the reality gap, sim-to-real transfer methods are extended to both visual \cite{james2017transferring, tobin2017domain, sadeghi2016cad2rl} and dynamics \cite{valassakis2020crossing, tsai2021droid, DBLP:journals/corr/abs-1710-06537,DBLP:journals/corr/RajeswaranGLR16, DBLP:journals/corr/YuLT17} solutions. The remainder of this section focuses on reviewing visual sim-to-real approaches.

\subsection{Domain Adaptation}

Domain adaptation is an approach to overcome the reality gap by learning transferable representations across different domains, either by minimising a distance measure between the two distributions \cite{long2015learning, tzeng2017adversarial,sun2016deep} or by learning pixel-level representations to make the simulated images as close as possible to their real-world counterparts \cite{bousmalis2018using, james2019sim}. While domain adaptation showed promising results in the literature, incorporating real-world images is necessary for successful transfers. In this work, we focus on the zero-shot sim-to-real transfer solutions that do not require finetuning with real-world images.

\subsection{Domain Randomisation}
One of the most common forms of sim-to-real transfer is domain randomisation, a simple-yet-promising technique to bridge the reality gap. Instead of collecting the training data from a single simulated environment, the model is exposed to a variety of random environments to enforce domain invariance. One of the earliest successes of domain randomisation is the pioneering work presented in \cite{sadeghi2016cad2rl}, where the authors leveraged the technique to train for a collision-free indoor flight. They showed that a policy trained solely in simulation is capable of adapting to real-world scenarios if exposed to a sufficient amount of randomness at training time. Several other works have then extended on this success for a variety of robotics manipulation tasks \cite{tobin2017domain, james2017transferring, andrychowicz2020learning, DBLP:journals/corr/abs-1712-07642, DBLP:journals/corr/abs-1806-07851, DBLP:journals/corr/abs-1810-03237, DBLP:journals/corr/abs-1709-07857, DBLP:journals/corr/abs-1710-06542, DBLP:journals/corr/abs-1810-05687}.

In \cite{james2017transferring}, the researchers investigated the use of domain randomisation with a task that requires hand-eye coordination (robotic grasping task). Their end-to-end approach succeeded in transferring to clear environments but failed to generalise to more cluttered ones. In \cite{tremblay2018training}, the focus was on training object detector using domain randomisation. Unlike other researches, they introduced the use of a non-realistic noise in the randomised scenes. Although their zero-shot model produced adequate results, they needed to incorporate real-world data for better performance. The authors in \cite{andrychowicz2020learning} exploited the randomisation of both physical and visual properties to train a dexterous robotic hand to perform an in-hand manipulation, where they further improved their work in \cite{akkaya2019solving} by proposing an iterative approach to learn the randomisation parameters' distributions, the so-called Automatic Domain Randomisation \cite{DBLP:journals/corr/abs-1907-01879, ruiz2019learning, heindl2020blendtorch}.

The authors in \cite{tobin2017domain} employed domain randomisation for the task of 2D object detection. Their model achieved a pose estimation error of $1.5$ cm upon tested in the real world. Their work, however, was tailored to regressing to the object position relative to the world frame, but not its orientation, which is crucial for the majority of robotics manipulation tasks. 

In general, the main goal of domain randomisation is to find what type of randomisation is effective in different scenarios. Both \cite{james2017transferring} and \cite{tobin2017domain} found that models are sensitive to the presence of distractors and textures in the simulated scenes. Several other factors have also been investigated in the literature, such as object textures \cite{james2017transferring,tobin2017domain, tremblay2018training, Loquercio_2020,8968139}, addition of distractors \cite{james2017transferring, tobin2017domain, tremblay2018training, Loquercio_2020}, lights properties \cite{tremblay2018training, Loquercio_2020}, target object shapes \cite{tobin2018domain, Loquercio_2020, 8968139}, and sample size \cite{tobin2017domain, james2017transferring, 8968139}. However, none of these works have benchmarked the different choices that must be made to achieve the optimal transfer. In this paper, we empirically evaluate two modes of design choices on a real-world 6D pose estimation task, which serves as a reasonable proxy for evaluating end-to-end control.

\section{Method}
This work is composed of two distinct experiments that are tailored to different objectives. In this section, we describe the setting shared between both experiments.

\subsection{Problem Definition}
The goal of the experiments is to train two models, $p_{O}(I_C)$ and $q_{O}(I_C)$, to map an RGB image $I$ captured from a camera $C$ to the 3D position $(x^{p}_{O},y^{p}_{O},z^{p}_{O})$ of the target object $O$ and its 4D quaternion $(x^{q}_{O}, y^{q}_{O}, z^{q}_{O}, w^{q}_{O})$ relative to the camera frame. We chose to regress to quaternions due to their compact form, although there are other possible representations for 3D rotations \cite{zhou2020continuity}. While we could train a single network to regress to the full 6D pose of the target object, as is often done in practice \cite{DBLP:journals/corr/KendallGC15}, we trained two separate networks to avoid the laborious process of finding the optimal balancing between the position and the orientation terms in the loss function, which would distract us from the main point of the paper. The training images are rendered in simulation, and depending on the experiment, they can be of different quality levels, where distractors might also be present in the environment. At test time, the trained model is used without any additional finetuning with real-world images.

\subsection{Training (Synthetic) Data Collection}
\label{training_data_collection}
We used Unity3D \cite{unity} to render all of our training images. We chose to work with Unity as it provides several render pipelines that are used to generate images of different quality levels. Within each sample scene, we define a plane upon which the target object is positioned. Further, we have a camera that moves and rotates in all the three dimensions to capture the target object in different poses, along with some directional lights to illuminate the scene. Depending on the experiment, we randomise different parameters while collecting the data to provide enough variability in the training dataset. For all experiments, however, the following aspects are randomised for each generated example:

\begin{itemize}
    \item The colour of the target object and its background
    \item The position, orientation, field of view, and focal length of the camera
    \item The number of lights, their positions, orientations, and specular characteristics
\end{itemize}

Table~\ref{tab:apendix_random_params} shows the full list of the randomisation parameters along with their values. The additional experiment-specific parameters are discussed in their respective sections separately. 

\begin{table}[!tb]
\caption{Randomisation parameters values.}
\label{tab:apendix_random_params}
\centering
\begin{tabular}{|m{2.6cm}|m{5.2cm}|}
\hline \xrowht[()]{5pt}
\textbf{Parameter}  & \textbf{Range of values}  \\ \hline 
Camera position & Placed within a box of size ($70\times40\times50$) cm around the target object \\ \hline 
Camera rotation & \begin{tabular}[c]{@{}l@{}}$x$: {[}\ang{0}- \ang{67.5}{]}\\ $y$: {[} \ang{-25}, \ang{75}{]} \\ $z$: {[} \ang{-10}, \ang{55}{]}\end{tabular} \\ \hline 
Camera field of view & $\pm$ \ang{5} from the estimated real-world equivalent \\ \hline 
Target object colour& $\pm$ 10\% from the estimated real-world equivalent   \\ \hline 
Number of lights & {[}0,3{]} directional lights, depending on the quality level  \\ \hline 
Lights intensity & \begin{tabular}[c]{@{}l@{}}HDRP: {[}0, 2000{]} lux\\ Built-in renderer: {[}0,1{]}\end{tabular} \\ \hline 
Lights position  & Placed within a box of size ($6\times2.8\times6$) m around the target object   \\ \hline 
Lights rotation   & \begin{tabular}[c]{@{}l@{}}$x$: {[}\ang{5}, \ang{30}{]}\\ $y$: {[}\ang{-180}, \ang{180}{]} \\ $z$: {[}\ang{-90}, \ang{90}{]}\end{tabular}   \\ \hline
\end{tabular}

\end{table}

\begin{figure}[tb] 
    \centering
    \stackunder[4pt]{%
        \includegraphics[width=2.1cm]{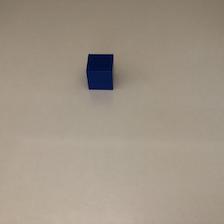}%
        }{\footnotesize Cube}%
    \stackunder[4pt]{%
        \includegraphics[width=2.1cm]{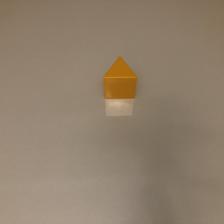}%
        }{\footnotesize Triangular Prism}%
    \stackunder[4pt]{%
        \includegraphics[width=2.1cm]{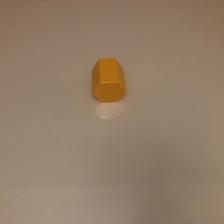}%
        }{\footnotesize Hexagon}%
    \stackunder[4pt]{%
        \includegraphics[width=2.1cm]{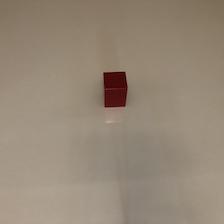}%
        }{\footnotesize Rectangular Prism}%
    \caption{The four primitive shapes that we used throughout the experiments.}
    \label{geometrical_shapes}
\end{figure}

\subsection{Testing (Real-world) Data Collection}

To address the research questions, we created mesh representations for four primitive shapes, shown in Fig.~\ref{geometrical_shapes}, that we used to evaluate both experiments. To find the accurate ground-truth values, we used \textit{ArUco} \cite{romero2018speeded}, an accurate marker-based pose estimation algorithm. For each test example, we placed a marker on top of the target object and carefully validated, with an accurate ruler, the offset from the object's coordinate frame. The image is then supplied to \textit{ArUco} which outputs the 6D pose value of the object relative to the calibrated camera. Another image is captured for the same scene without the marker to be used as an input to the trained networks.

\begin{figure}[!tb] 
    \centering
      \includegraphics[width=8.5cm]{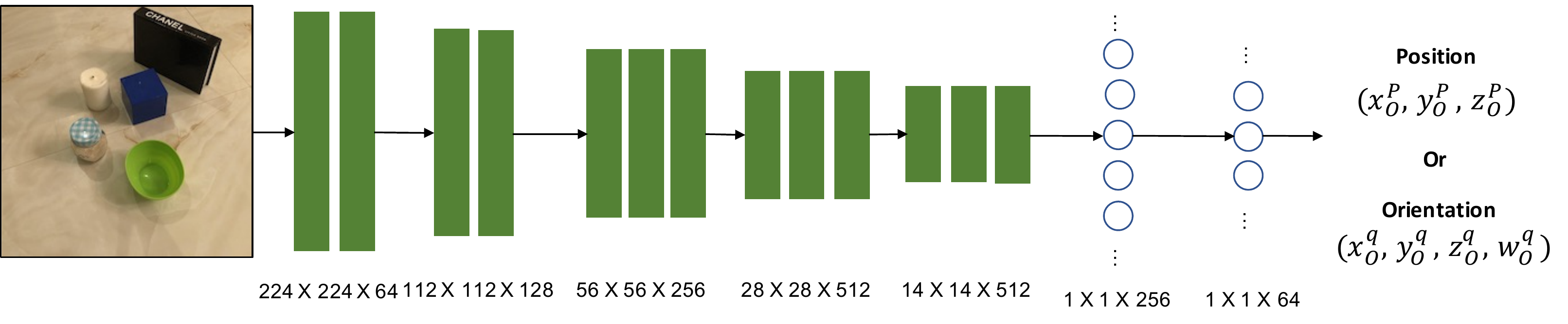}
    \caption{The end-to-end task of 6D pose estimation. For each target object $O$, we train two separate networks, $p_{O}(I_C)$ and $q_{O}(I_C)$, that map a single image to the 6D pose of $O$. Each input image is downsized to ($224 \times 224$) and supplied to two separate networks to be mapped to a tuple of $(x^{p}_{O},y^{p}_{O},z^{p}_{O})$ that represents the 3D position of the target object and a tuple of $(x^{q}_{O}, y^{q}_{O}, z^{q}_{O}, w^{q}_{O})$ which represents the target's 3D orientation.}
    \label{fig:network_arch}
\end{figure}

\subsection{Network Architecture}

For all experiments, we used a modified version of the VGG-16 \cite{simonyan2014deep} network architecture, which was proposed in \cite{tobin2017domain}. As shown in Fig.~\ref{fig:network_arch}, the network consists of five groups of convolutional layers each with a kernel size of $3\times3$ and a stride of $1$, where dimensionality reduction is performed after each group. To guarantee a fair comparison, we fixed the hyper-parameters, the seed, and the loss function for all experiments. All models are trained using \textit{Adam} optimiser until convergence with a starting learning rate of $1e-4$ and a scheduler to decrease it by a factor of $0.1$ every $30$ epochs. The networks weights are initialised randomly, and both the inputs and outputs are normalised before starting the training process.

\subsection{Evaluation Metrics}
To evaluate the performance of our models, we test them using the collected real-world images and compare the actual labels to the results obtained from the networks. The position error is computed using the Root Mean Squared Error (or RMSE) formula. For the orientation, however, we use Equation~\ref{eq:quat-error} \cite{mahendran20173d} to find the angle required to rotate from one quaternion (the label, $q_i$) to the other (the prediction, $\hat{q_i}$).

\begin{equation} \label{eq:quat-error}
    \theta_i = 2\times \cos^{-1}({\left|\langle \hat{q_i}, q_i \rangle\right|}) 
\end{equation}
    
Where $\langle \hat{q_i}, q_i \rangle$ is the inner product between the label and the prediction of the $i^{th}$ sample image.

\section{Experiment 1: Rendering Quality}
This experiment aims at studying the impact of the simulator's fidelity on the overall model's performance in the real world. More precisely, our focus is on finding answers to the following three questions:
\begin{enumerate}
    \item How critical is the quality of the simulator for achieving successful sim-to-real transfers?
    \item What is the effect that each simulation parameter has on the overall sim-to-real transfer performance?
    \item What is the optimal trade-off of low-quality and high-quality images given a fixed amount of rendering time?
\end{enumerate}

\begin{table*}[tb]\centering\setlength\tabcolsep{2.5pt}\renewcommand\arraystretch{1.25}
\centering
 \caption{An overall comparison between all the considered levels of simulation.}
 \label{tab:comparison}
  \noindent\makebox{%
    \begin{tabular}{|p{4cm}|p{2.4cm}|p{2.3cm}|p{2.3cm}|p{2.3cm}|p{2.3cm}|}
      \hline
      \diagbox[width=4cm, height=0.7cm]{ Quality level }{Feature} &Directional Lights&Shadows&Anti-aliasing&Dithering effect&Render Pipeline \\
      \hline
      1  & No & No & No & No & \multirow{3}{*}{Built-In}\\ \cline{1-5} 
      2  & Yes & No & No & No & \\  \cline{1-5} 
      3  & Yes & Low-quality & No & No & \\ 
      \hline
      4  & Yes & Medium-quality & No & No & \multirow{5}{*}{HDRP}\\  \cline{1-5} 
      5  & Yes & Medium-quality & Low-quality & No & \\  \cline{1-5}
      6  & Yes & High-quality & Low-quality & No & \\  \cline{1-5} 
      7  & Yes & High-quality & High-quality & No &  \\  \cline{1-5} 
      8  & Yes & High-quality & High-quality & Yes &  \\  
      \hline
    \end{tabular}
  }
\end{table*}

\begin{figure*}[!tb]
    \centering
    \includegraphics[width=0.95\linewidth]{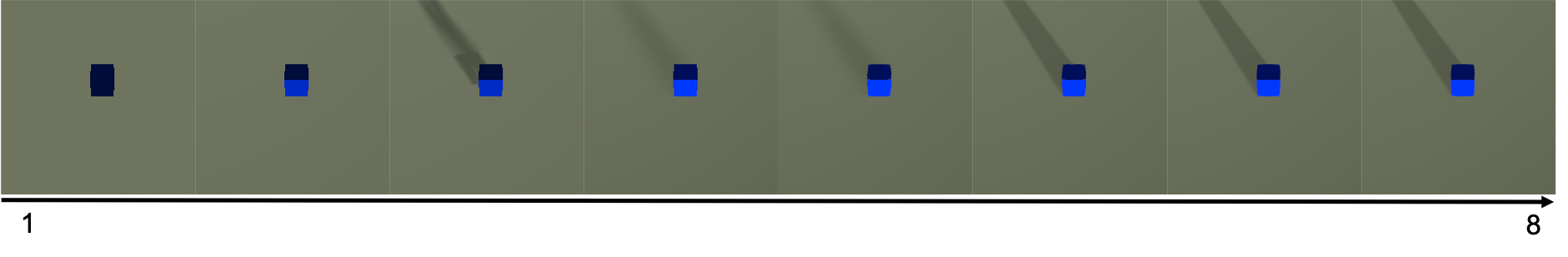}
    \caption{Sample images from each quality level. From left to right: low-quality (level 1) to high-quality (level 8).}
    \label{fig:qualitylevels}
\end{figure*}

\subsection{Experimental Setup}

To achieve this, we use two renderers provided by Unity: the Built-in and the High Definition Render Pipeline (HDRP hereafter) \cite{unity3d-comparison-hdrp} to render images of eight quality levels. As presented in Table~\ref{tab:comparison}, these levels vary in terms of the following aspects:

\begin{itemize}
    \item \textbf{Directional lights}: we enable them for all quality levels except the baseline (level 1).
    \item \textbf{Shadows}: we alter the quality of the shadows for each simulation level to range from low to high quality. 
    \item \textbf{Anti-aliasing}: anti-aliasing is a technique used to overcome the staircase effect that usually occurs in physical simulators by smoothing the object boundaries and corners. 
    \item \textbf{Dithering effect}: dithering is a form of noise that is intentionally applied to rendered images to eliminate unwanted graphical issues, such as colour banding. 
    \item \textbf{Render pipeline}: the renderer used to generate the dataset.
\end{itemize}

In Fig.~\ref{fig:qualitylevels}, we visually illustrate the differences between the eight simulation quality levels. In this experiment, we assumed that the colours of both the target object and its background are known at training time, and no distractors are present in the environment. For each object, we collected eight training datasets for the different quality levels we consider, each with $30,000$ randomised synthetic images.  We tested the models with $100$ real-world testing images, per object, that are collected with a variety of poses and lighting conditions.

\begin{table}[!tb]
\caption{The average error across four target objects.}
\label{quality_rendering_results}
\begin{center}
\begin{tabular}{|m{2cm}|m{1.5cm}|m{1.5cm}|m{1.5cm}|}
\hline 
 Quality level & $xy$-position (cm) & $z$-position (cm) & Orientation (degrees) \\ 
\hline 
1 & $4.37 \pm 4.45$  & $12.46 \pm 8.06$ & $13.0 \pm 4.72$ \\
\hline 
2  & $1.44 \pm 1.68$  & $3.44 \pm 2.87$ & $4.57 \pm 3.38$ \\ 
\hline 
3 & $1.40 \pm 1.53$  & $3.22 \pm 3.05$ & $4.14 \pm 2.97$ \\
\hline 
4 & $0.83 \pm 1.10$  & $2.73 \pm 3.60$ & $3.52 \pm 2.26$ \\
\hline 
5 & $0.70 \pm 0.62$  & $2.67 \pm 2.22$ & $3.34 \pm 2.41$ \\
\hline 
6 & $0.63 \pm 0.63$  & $2.50 \pm 2.14$ & $3.26 \pm 2.40$ \\
\hline 
7 & $0.61 \pm 0.62$  & $2.43 \pm 2.07$ & $3.20 \pm 2.57$ \\
\hline
8 & $\mathbf{0.53 \pm 0.52}$  & $\mathbf{2.38 \pm 1.93}$ & $\mathbf{3.08 \pm 2.15}$ \\
\hline 
\end{tabular}
\end{center}
\end{table}

\subsection{Results}

In Table~\ref{quality_rendering_results}, we present the average error achieved by each quality level across the four target objects. We reported the $xy$ and the $z$ position errors separately to study the effect on the different tasks independently: $xy$ position (across the image plane), $z$ position (perpendicular to the image plane), and 3D orientation.

\textbf{Altering the Rendering Quality}\quad We can conclude from Table~\ref{quality_rendering_results} that, as we increase the quality of the training data, the models performed consistently better, which answers our first question.

\textbf{Directional Lights Effect}\quad As we can deduce from the first two rows of Table~\ref{quality_rendering_results}, we were able to reduce the error by $\approx 68.4\%$ after enabling directional lights, which proves their importance for achieving successful transfer.

\textbf{Shadows Effect}\quad  Without shadows, as demonstrated in the second and third rows of Table~\ref{quality_rendering_results}, the models performed the worst when tested in the real world. Moreover, as we increased the quality of the shadows (level 3 vs. level 4), the performance of the models has significantly improved, which indicates the importance of the quality of the shadows.

\begin{figure*}[!tb]
    \centering
    \subfloat[$xy$-position]{\includegraphics[width=0.29\linewidth]{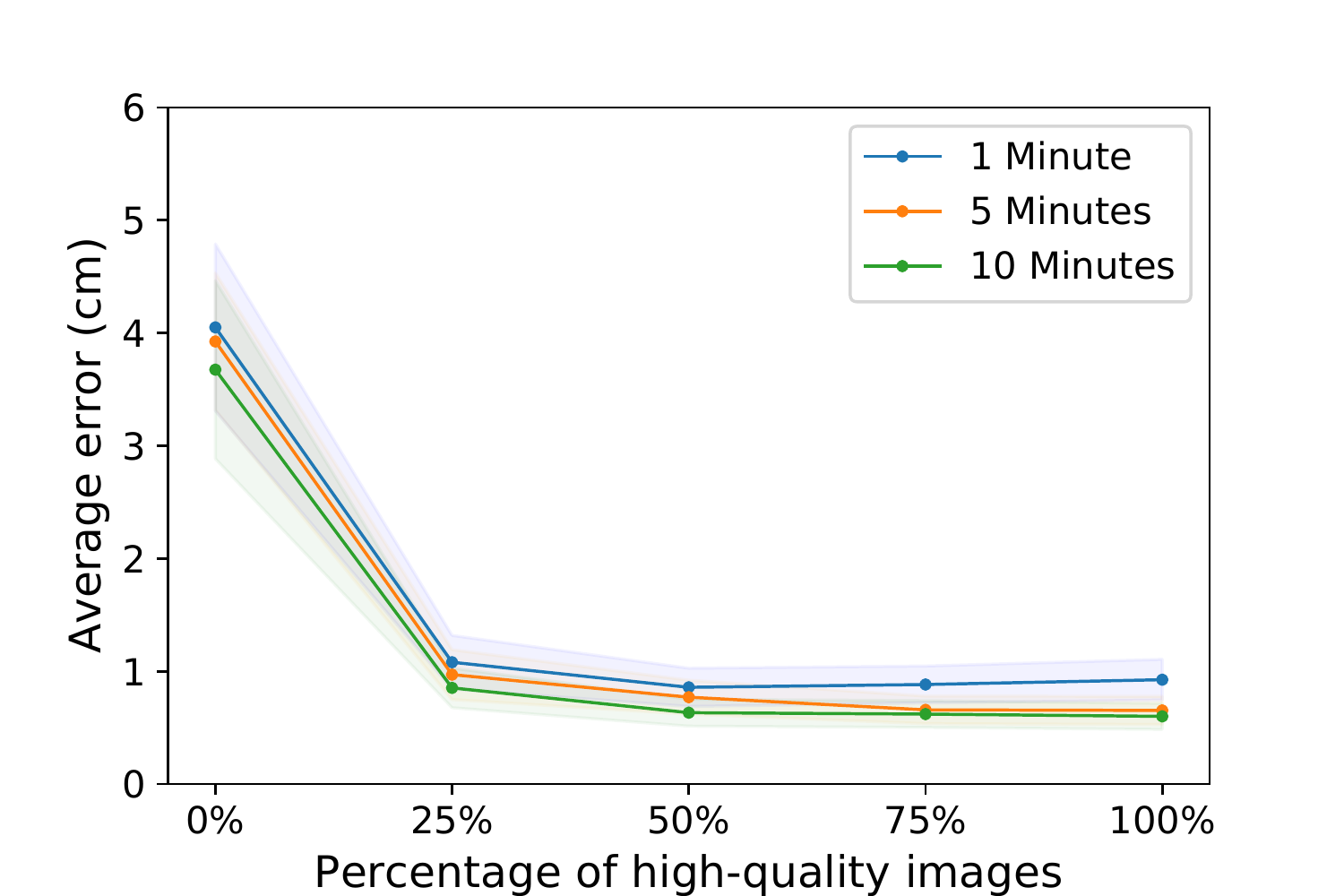}} 
    \hspace{1em}
    \subfloat[$z$-position]{\includegraphics[width=0.29\linewidth]{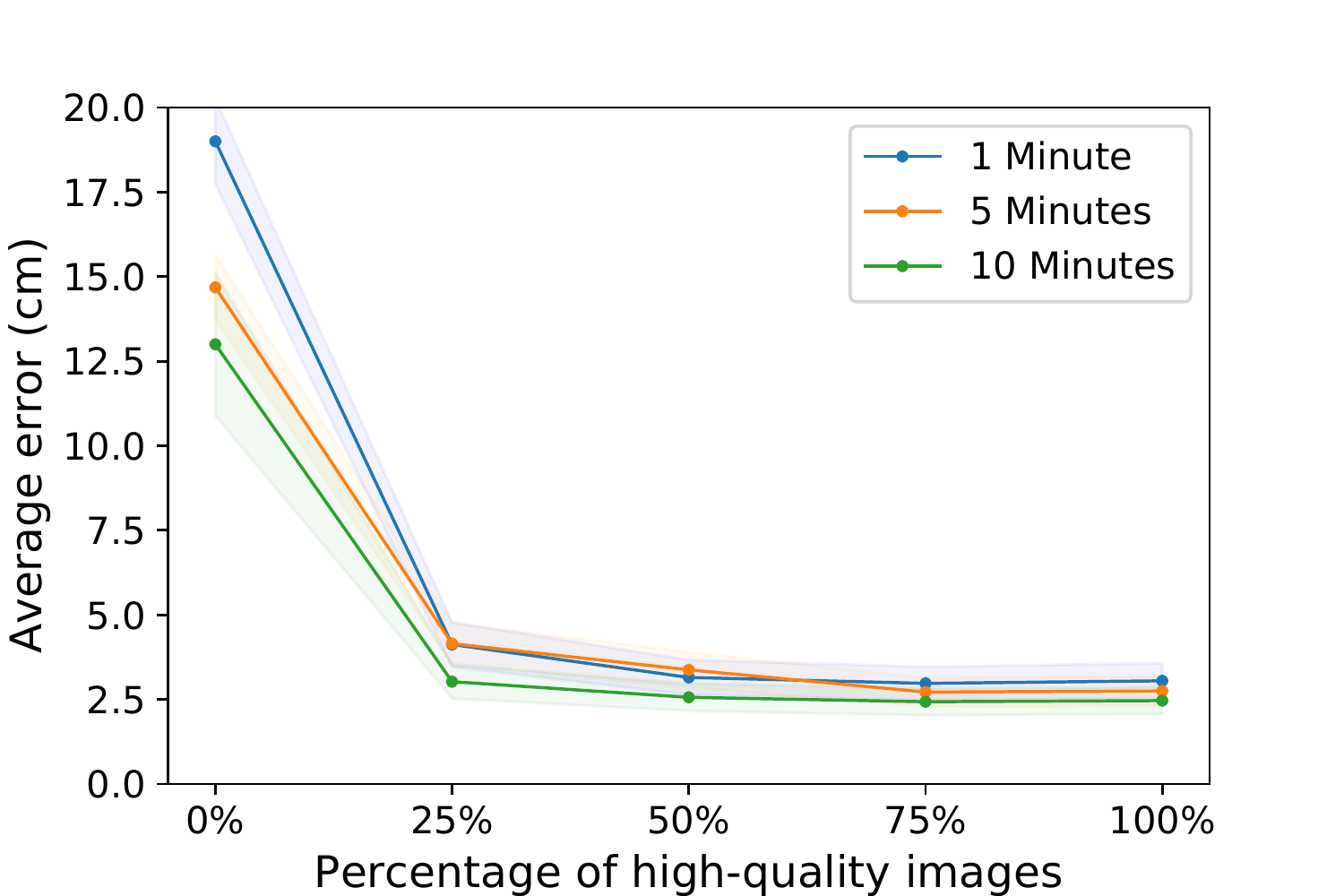}} 
    \hspace{1em}
    \subfloat[Orientation]{\includegraphics[width=0.29\linewidth]{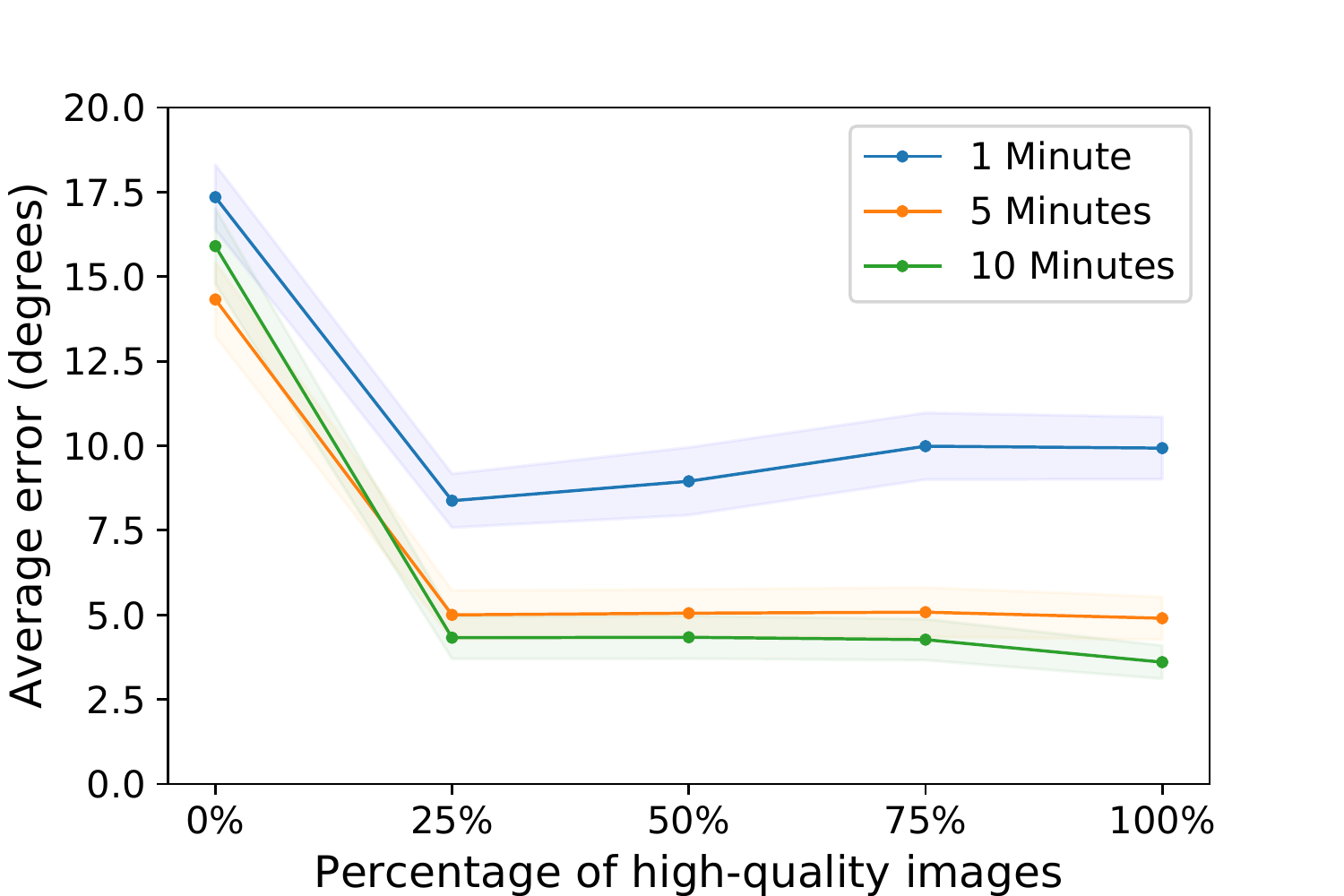}}
    \caption{The results obtained after training the models with combinations of high-quality and low-quality training images, under three different data collection time constraints.}
    \label{fig:comb_error}
\end{figure*}

\textbf{Anti-aliasing Effect} \quad Recall from Table~\ref{tab:comparison} that the only difference between level 4 and level 5 images is the use of anti-aliasing. As we can conclude from Table~\ref{quality_rendering_results}, models trained with level 5 images succeeded in outperforming the ones trained with no anti-aliasing enabled (i.e. level 4), which shows the positive impact of anti-aliasing on the overall models' performance. However, since the anti-aliasing used with level 5 images is of low-quality, it might fail, for some objects with complex shapes, to fix the staircase issues.

\textbf{Dithering Effect} \quad The last two rows of Table~\ref{quality_rendering_results} demonstrate the results before and after enabling this effect, where we can see that dithering plays a vital role in boosting the transfer performance.

\subsection{Combining High-quality and Low-quality Images}
In the previous set of experiments, we showed that models trained with high-quality images have superior performance compared to the ones trained with lower-quality levels. Nonetheless, rendering high-quality images comes with an additional computational cost. We found with experiments that rendering one image from level 1 takes an average of only $8.4$ ms, while generating one high-quality image (level 8) requires more than triple the time, with an average of $27.2$ ms. Rather than training with images of only one level of quality, we could also train with both high-quality and low-quality images, to combine the merits of both fast and accurate rendering.

Fig.~\ref{fig:comb_error} shows the results obtained after training several models under different time constraints. For each experiment, we fixed the data collection time and varied the percentage of each type of images. As we can deduce from the graphs, the overall real-world test error dropped significantly after incorporating as minimum as $25\%$ of high-quality images. Surprisingly, however, as we increased the percentage of high-quality images, the performance of the models exhibited no clear trend, as it sometimes stabilised, reduced, or improved slightly. The overall conclusion is therefore that if a significant number of images can be collected (i.e. no time constraint), there is no harm from rendering all images with high quality. Otherwise, if only a small number of images can be collected, including some high-quality ones (e.g. 25\%) should significantly improve the results.

\section{Experiment 2: Randomisation Type}

\begin{figure}[!tb] 
    \centering
    \includegraphics[width=8cm]{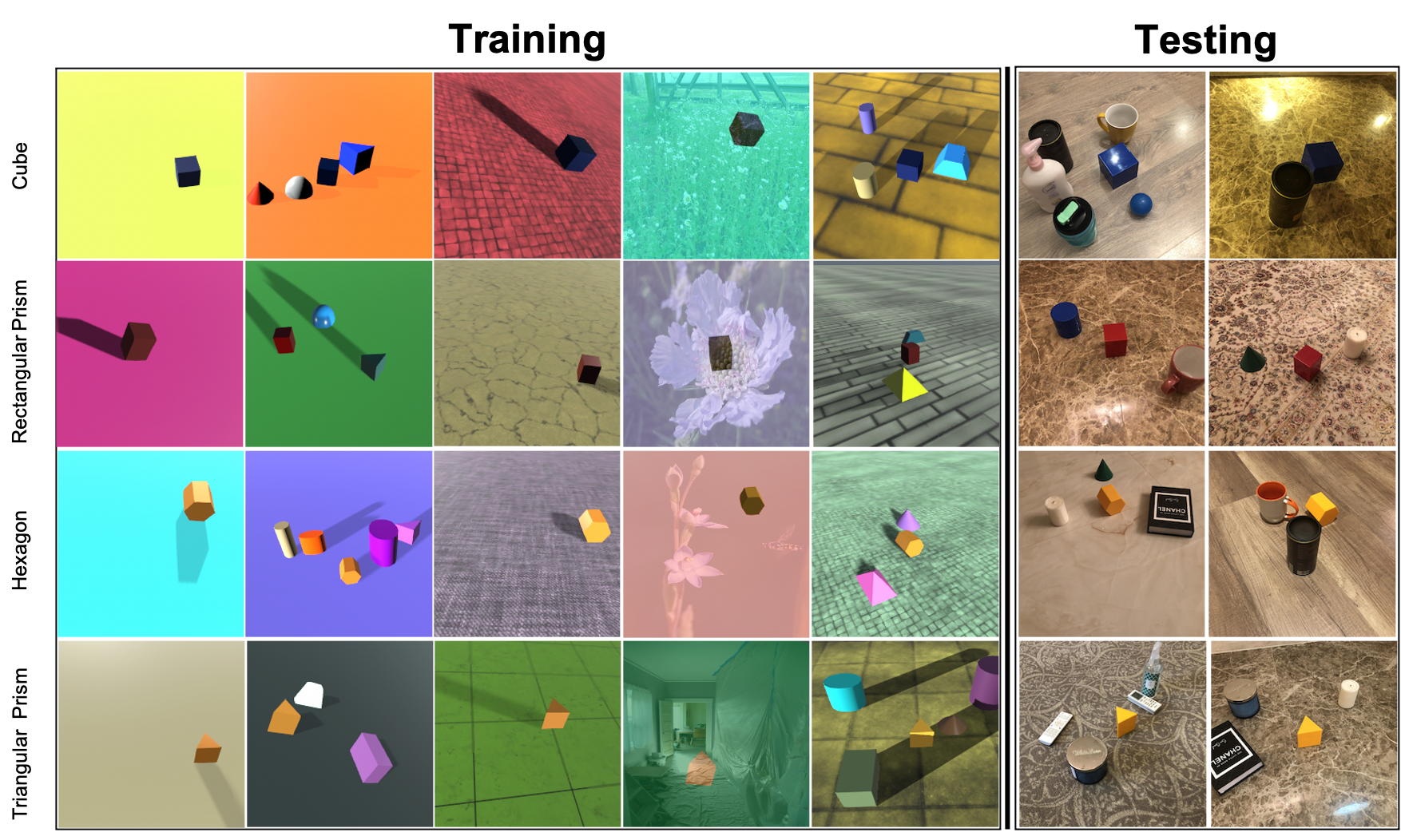}%
    \caption{Left: high-quality synthetic images collected with random colours, textures and distractors. Right: real-world testing images labelled using ArUco. }
    \label{sampletraintest}
\end{figure}


\begin{figure}[!tb] 
    \centering
    \subfloat[Baseline \protect\\(colours only)]{%
        \includegraphics[width=2cm]{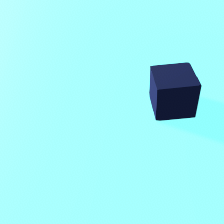}%
        }%
    \hspace{1em}%
     \subfloat[Baseline \protect\\+ ImageNet \protect\\background]{%
        \includegraphics[width=2cm]{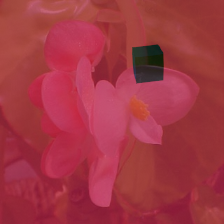}%
        }   
    \hspace{1em}%
    \subfloat[Baseline \protect\\+ textures]{%
        \includegraphics[width=2cm]{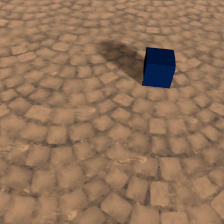}%
        }
    \hspace{1em}%
    \subfloat[Baseline \protect\\+ distractors]{%
        \includegraphics[width=2cm]{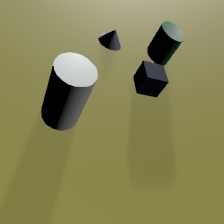}%
        }%
    \hspace{1em}%
     \subfloat[Baseline \protect\\+ textures \protect\\+ distractors]{%
        \includegraphics[width=2cm]{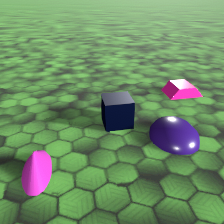}%
        }  
    \caption{A sample from each of the five datasets collected to study the impact of the different randomisation settings on the sim-to-real transfer performance. In these examples, the target object is the blue cube.}
    \label{exp2_example_images}
\end{figure}

\begin{figure*}[!tb]
    \centering
    \subfloat[$xy$-position]{\includegraphics[width=0.29\linewidth]{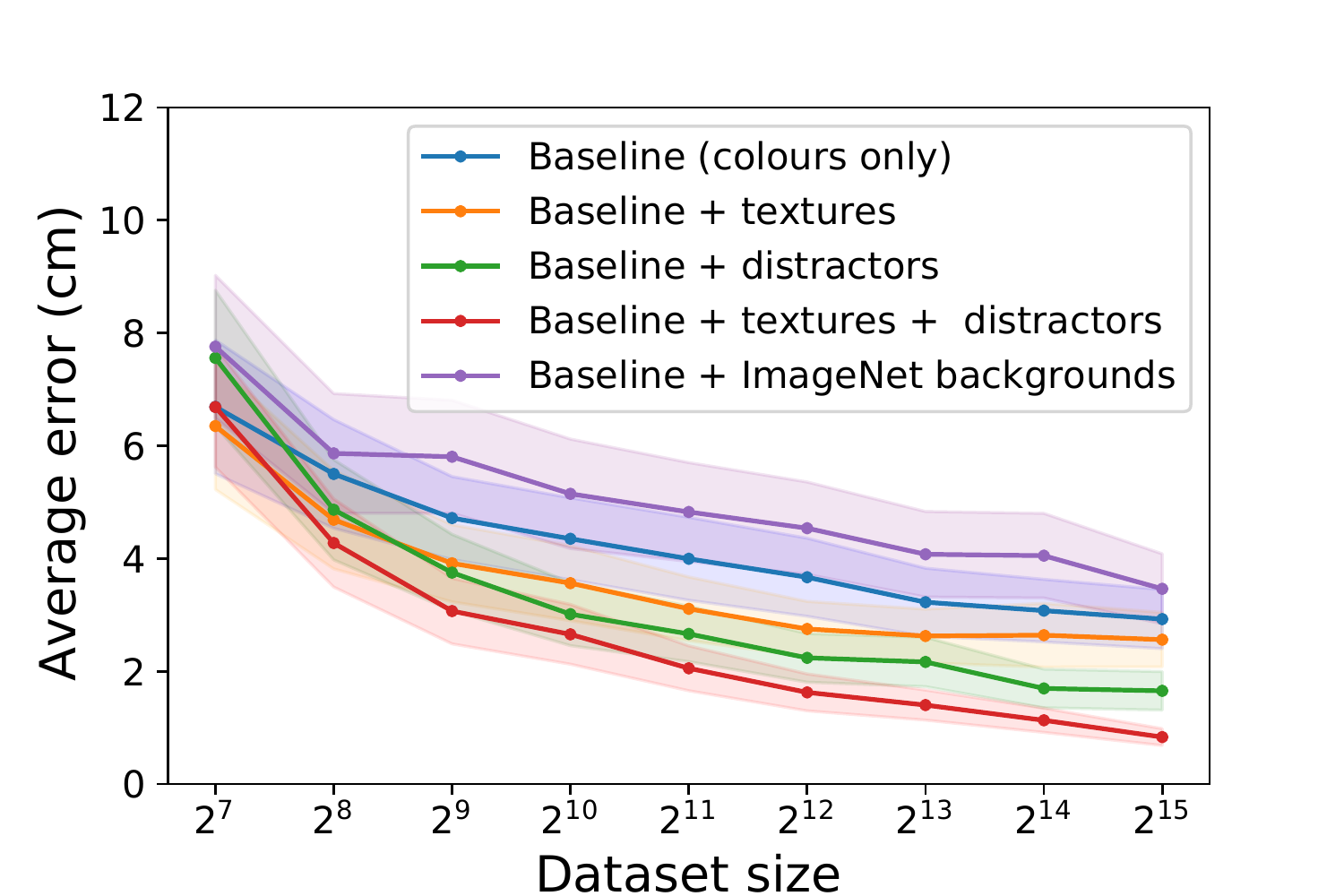}} 
    \hspace{1em}
    \subfloat[$z$-position]{\includegraphics[width=0.29\linewidth]{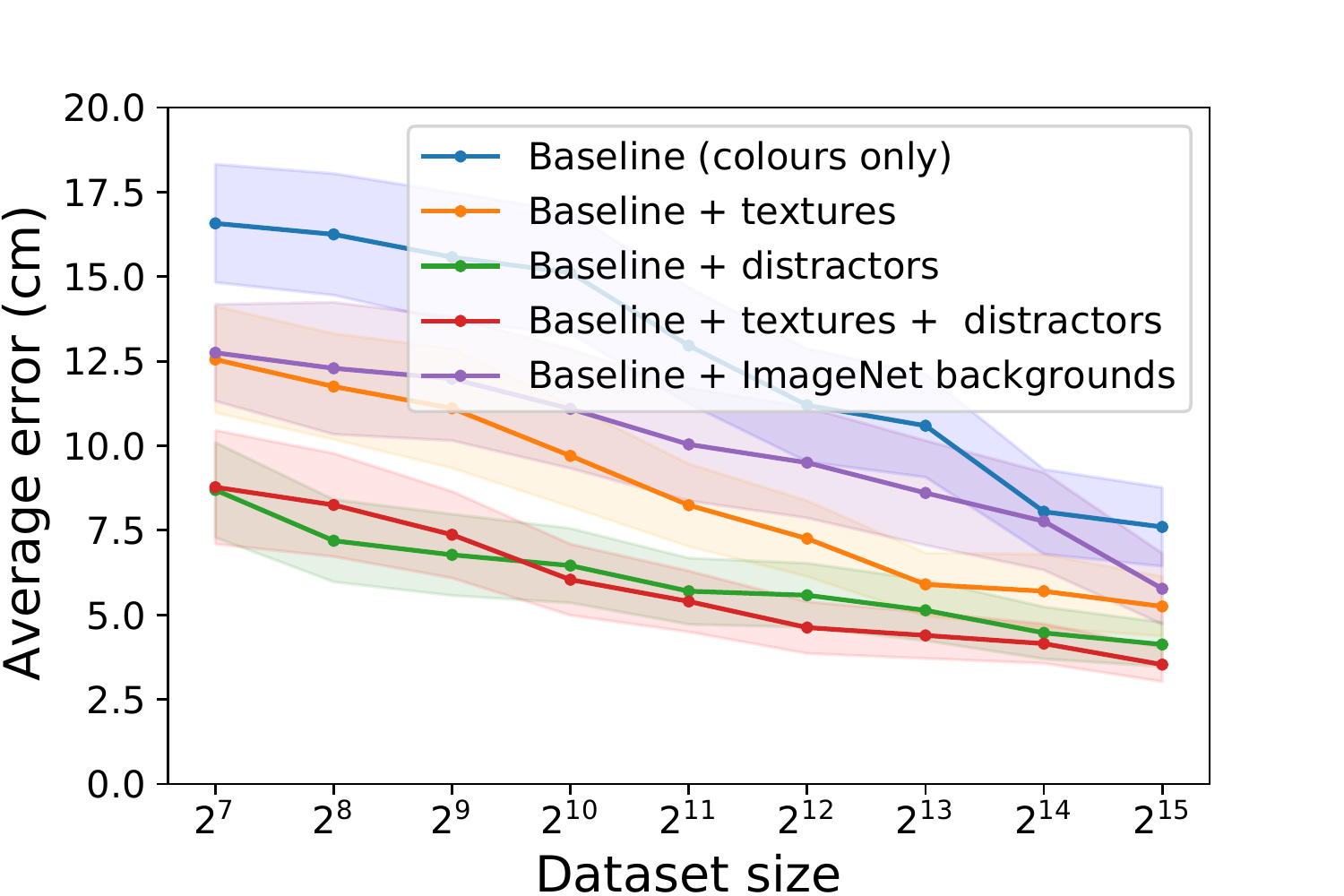}}
    \hspace{1em}
    \subfloat[Orientation]{\includegraphics[width=0.29\linewidth]{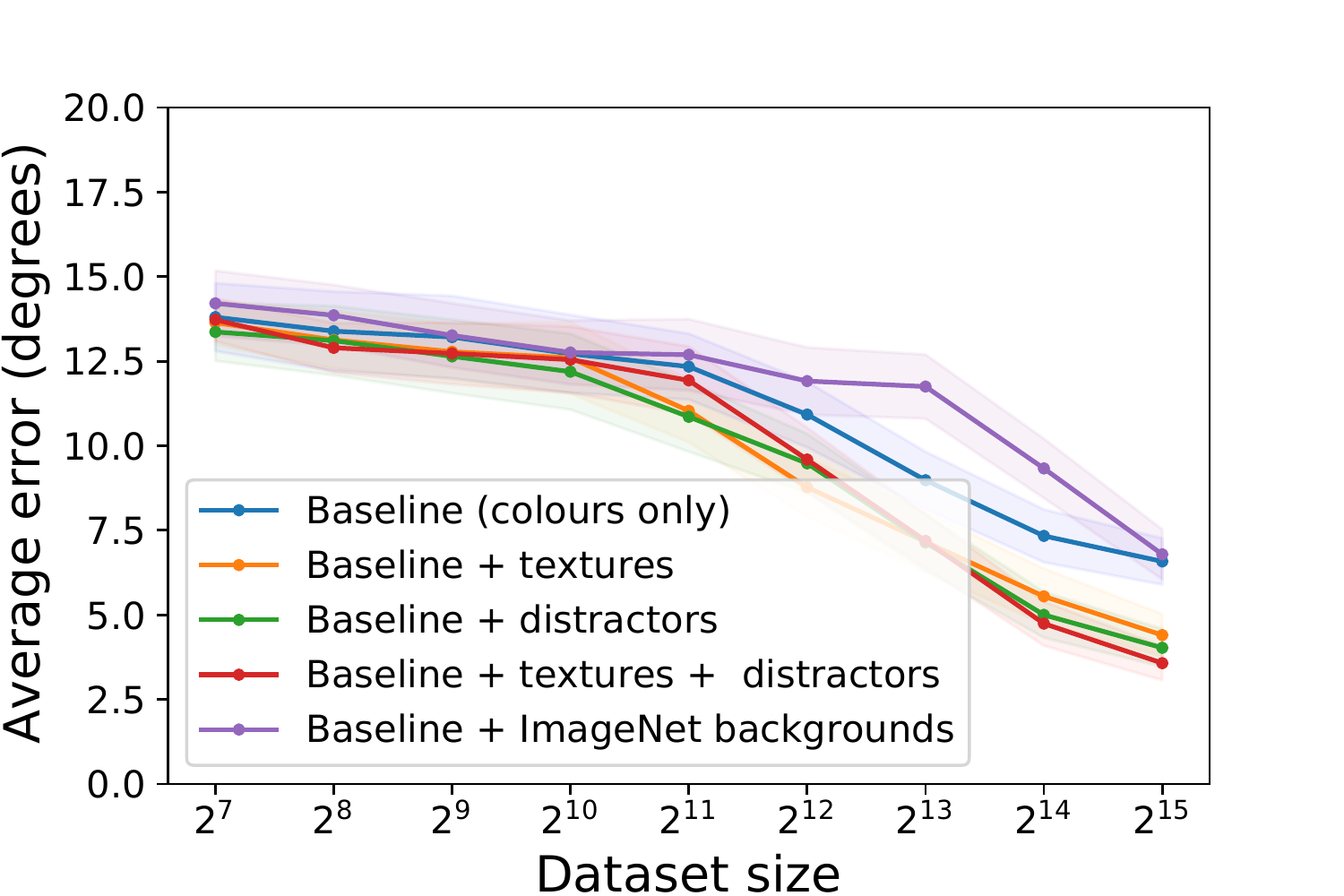}}
    \caption{The sensitivity of the models to distractors, textures, and number of images. The results are averaged across the four target objects.}
    \label{fig:exp2_results}
\end{figure*}

In this experiment, we examine the performance of models trained with high-quality synthetic images, when tested in varied and cluttered real-world environments. More precisely, we strive to assess the significance of different randomisation settings to the models' transferability to the real world. Similar to the previous experiment, the focus here is on evaluating the sim-to-real performance, rather than studying the different settings of 6D pose estimation. 

\subsection{Experimental Setup}

Unlike the previous experiment, in this, we assumed that the models have no prior knowledge about the background of the target object (i.e. it can be of any texture or colour). In addition, distractors of any shape, colour, and size are presented at test time.

A summary of the approach is provided in Fig.~\ref{sampletraintest}. The models were trained using high-quality simulated images (left) and evaluated in the real-world (right). Random geometrical shapes were used as distractors at training time. However, a variety of never-seen distractors are presented in the real-world evaluation images, e.g. a cup, book, and candle. To assess which types of randomisation are most crucial for achieving successful transfer, we collected five training datasets, shown in Fig.~\ref{exp2_example_images}, which have the following specifications:

\begin{enumerate}
    \item Baseline: images with random solid colours in the background, but no textures or distractors
    \item Baseline + random ImageNet backgrounds
    \item Baseline + random background textures
    \item Baseline + random distractors
    \item Baseline + random textures and distractors
\end{enumerate}

The colour of the target object was uniformly sampled around our best estimate of its colour in the real world. All the five datasets were collected with high-quality anti-aliasing, high-quality shadows, and dithering effect. For each target object, we collected $110$ real-world testing images with a variety of backgrounds and distractors. We did not control for lighting and shadows in the real world, meaning that some images were captured from relatively dark scenes. For this experiment, we focused on answering the following three questions:

\begin{figure}[!tb]
    \centering
    \subfloat{\includegraphics[width=2.3cm]{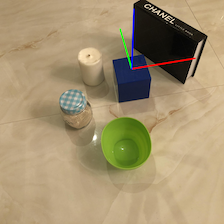}}
    \hspace{0.3em}%
    \subfloat{\includegraphics[width=2.3cm]{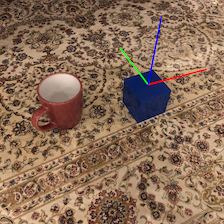}}
    \hspace{0.3em}%
    \subfloat{\includegraphics[width=2.3cm]{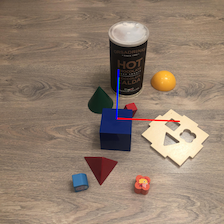}}
    \caption{Examples where the model succeeds in accurately estimating the 6D pose of the target object (cube). The axes represent the cube's local frame, and their colours correspondences are: $x$: red, $y$: green, and $z$: blue.}
    \label{successful_case}
\end{figure}

\begin{figure}[!tb]
    \centering
    \subfloat{\includegraphics[width=2.3cm]{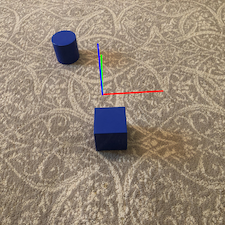}}
    \hspace{0.3em}%
    \subfloat{\includegraphics[width=2.3cm]{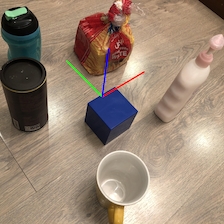}}
    \hspace{0.3em}%
    \subfloat{\includegraphics[width=2.3cm]{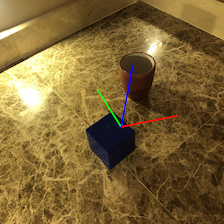}}
    \caption{Examples where the model fails to locate the target. The model missed the object when a distractor of the same colour is presented in the environment and with extremely cluttered or dark scenes. The axes are the cube's local frame ($x$: red, $y$: green, $z$: blue).}
    \label{fig:failure_modes}
\end{figure}

\begin{enumerate}
    \item What elements are most important to randomise while collecting the training data?
    \item How does the performance of the models change as we increase the training dataset size?
    \item How robust are the trained models to never-seen backgrounds and distractors?
\end{enumerate}

\subsection{Results}

\textbf{Type of Randomisation} \quad We assessed the sensitivity of our models to both textures and distractors. We found, as shown in Fig.~\ref{fig:exp2_results}, that the models are sensitive to both factors for the three tasks. However, we noticed that for position estimation, training with distractors in the environment (the green line in Fig.~\ref{fig:exp2_results}) is of more importance than having textures, which is concluded from the fact that models trained with textures and no distractors had larger errors compared to the ones trained with distractors only. In general, both the baseline and the ImageNet datasets failed with the worst performance compared to the rest of the models. In contrast, the full dataset attained the best performance in all cases.

\textbf{Varying the Dataset Size} \quad To answer the second question, we trained several models on different dataset sizes, ranging from $2^7$ to $2^{15}$ images. The graphs in Fig.~\ref{fig:exp2_results} show that as we increase the training dataset size, the average real-world error gradually decreased for all the five datasets, as would be expected.

\textbf{Qualitative Results} \quad Overall, we observed that our models, when trained with both textures and distractors, are robust to changing environments and can generalise to cases with never-seen distractors and backgrounds. In Fig.~\ref{successful_case}, we show some examples when the model precisely estimated the 6D pose of the target object, in spite of the novel backgrounds and distractors. Despite the promising results, the model was not able to accurately estimate the pose of the target object, as shown in Fig.~\ref{fig:failure_modes},  when another distractor of the same colour is presented in the environment. Further, extremely cluttered environments and dark scenes posed a challenge to the model.

\section{Conclusion}
This paper investigated two different design choices in domain randomisation for visual sim-to-real transfer: the quality of the renderer, and the type of randomisation. For the first set of experiments, we found there to be a very strong relationship between the quality of the renderer, and the sim-to-real performance. We also proposed to combine low-quality images with high-quality images, and found that for the same overall rendering time, it is more important to have a high percentage of high-quality images, than low-quality images. For the second set of experiments, we showed that randomising both distractor objects and background textures are important for generalising to novel environments. These conclusions can now be used by others in designing their own domain randomisation datasets, with a view towards achieving optimal sim-to-real performance for a given amount of dataset generation time.

\bibliographystyle{IEEEtran}
\bibliography{references}

\end{document}